\documentclass[letterpaper,10pt,conference,final]{ieeeconf}

\IEEEoverridecommandlockouts   

\usepackage{times}
\usepackage[latin1]{inputenc}

\usepackage{amsmath,empheq,nicefrac,amssymb,bm,mathtools}%https://www.overleaf.com/project/6207b984662ceb38d8137618
\usepackage[super]{nth}

\usepackage[nocomma]{optidef}

\usepackage{multicol}

\usepackage{epsfig}

\usepackage{graphicx}
\usepackage[crop=pdfcrop,mode=batch]{pstool}

\usepackage{booktabs}
\usepackage{multirow}

\usepackage[sort,compress,space]{cite}

\usepackage{balance}

\usepackage{url}
\usepackage[bookmarks=true]{hyperref}
\definecolor{mygreen}{rgb}{0,0.6,0}
\definecolor{myblue}{rgb}{.12,.46,.70}
\hypersetup
{colorlinks=true,
	linkcolor=red,
	citecolor=mygreen,
	filecolor=magenta,
	urlcolor=myblue,
	linktoc=page,
}

\DeclareMathAlphabet{\pazocal}{OMS}{zplm}{m}{n}

\newcommand{\R}[1]{\ensuremath{\mathbb{R}^{#1}}}

\newcommand{\vr}[1]{{\mbox{\bm{$#1$}}}}

\newcommand{\trans}{^{\ensuremath{\mathsf{T}}}}

\newcommand{\dx}{\ensuremath{\vr{\dot{x}}}}
\newcommand{\dq}{\ensuremath{\vr{\dot{q}}}}
\newcommand{\dv}{\ensuremath{\vr{\dot{v}}}}
\newcommand{\ddq}{\ensuremath{\vr{\ddot{q}}}}

\title{
	\vspace{-15mm}
	\begin{center}
		{\small The final version of this paper was published in IROS'2022 (Kyoto)} \\ [-.6em]
		{\small DOI link: \url{https://doi.org/10.1109/IROS47612.2022.9981255}}     \\ [.6em]
		A Legendre-Gauss Pseudospectral Collocation Method                          \\ 
		for Trajectory Optimization in Second Order Systems
	\end{center}
}

\author{Siro Moreno-Mart\'{\i}n, Llu\'{\i}s Ros, and Enric Celaya% <-this % stops a space
	\thanks{This work has been partially supported by the Spanish Agencia Estatal de Investigaci\'on under the {\sc Kinodyn+} project, with reference PID2020-117509GB-I00/AEI/10.13039/50110001103. The authors are with the Institut de Rob\`otica i Inform\`atica Industrial, CSIC-UPC, Barcelona, Spain. Contact emails: {\tt\small \{smorenom,ros,ecelaya\}@iri.upc.edu}}%
}

\begin{document}
%-----------------------------------

\maketitle
\thispagestyle{empty}
\pagestyle{empty}

%-----------------------------------
\begin{abstract}
%-----------------------------------
%
% Versió editada pel Lluís
%
Pseudospectral collocation methods have proven to be powerful tools to solve optimal control problems. While these methods generally assume the dynamics is given in the first order form \mbox{$\dx = \vr{f}({\vr{x}, \vr{u}},t)$}, where $\vr{x}$ is the state and $\vr{u}$ is the control vector, robotic systems are typically governed by second order ODEs of the form $\ddq = \vr{g}(\vr{q}, \dq, \vr{u},t)$, where $\vr{q}$ is the configuration. To convert the second order ODE into a first order one, the usual approach is to introduce a velocity variable $\vr{v}$ and impose its coincidence with the time derivative of $\vr{q}$. Lobatto methods grant this constraint by construction, as their polynomials describing the trajectory for $\vr{v}$ are the time derivatives of those for $\vr{q}$, but the same cannot be said for the Gauss and Radau methods. This is problematic for such methods, as then they cannot guarantee that $\ddq = \vr{g}(\vr{q}, \dq, \vr{u},t)$ at the collocation points. On their negative side, Lobatto methods cannot be used to solve initial value problems, as given the values of $\vr{u}$ at the collocation points they generate an overconstrained system of equations for the states. In this paper, we propose a Legendre-Gauss collocation method that retains the advantages of the usual Lobatto, Gauss, and Radau methods, while avoiding their shortcomings. The collocation scheme we propose is applicable to solve initial value problems, preserves the consistency between the polynomials for $\vr{v}$ and $\vr{q}$, and ensures that $\ddq = \vr{g}(\vr{q}, \dq, \vr{u},t)$ at the collocation points.
\end{abstract}
%-----------------------------------

%%%%%%%%%%%%%%%%%%%%%%%%%%%%%%%%%%%%
\section{Introduction}
\label{sec:introduction}
%%%%%%%%%%%%%%%%%%%%%%%%%%%%%%%%%%%%

Several methods are available to solve optimal control problems in robotics. Among them, those based on direct collocation enjoy widespread adoption due to their advantages over indirect methods, which include their ease of use and their larger regions of convergence towards optimal solutions. Pseudospectral methods, also known as orthogonal methods, are a particular type of direct collocation methods that are attractive because of their exponential convergence properties, and they have been successfully applied to a variety of problems, including motion planning of robot arms \cite{zhao2018TrajOpt}, biped gait generation \cite{hereid2016biped}, contact implicit optimization \cite{patel2019contact}, or maneuver planning on the International Space Station~\cite{Bedrossian2007ISS}. Substantial theoretical work has also been done in pseudospectral optimal control theory~\cite{ross2012pseudospectral}. 

A pseudospectral method approximates each component of the state and control vectors using high degree polynomials, and imposing the dynamic equations at a set of collocation points. Depending on the method, these points are taken from the roots of specific orthogonal polynomials, or of combinations of such polynomials and their derivatives. The most common sets of collocation points are those based on Legendre or Chebyshev orthogonal polynomials, and depending on whether they include the bounds of the time interval or not, they are classified into Lobatto points, which include the two bounds of the time interval, Radau points, which include only one bound, and Gauss points, which include none. According to this classification, the most usual methods found in the literature are the Legendre-Gauss (LG), Legendre-Gauss-Radau (LGR), Legendre-Gauss-Lobatto (LGL), and Chebyshev-Lobatto (CHL) methods~\cite{Elnagar1998,garg2010unified}.
%The important difference between these sets of points is that the Lobatto points include the two bounds of the time interval, the Radau points include only one bound, and the Gauss point includes none. 

For problems with non-smooth solutions it could be convenient to partition the time domain into subintervals and use a different polynomial for each subinterval, but in this work we assume that a single global polynomial is used for the whole time domain. As is common in pseudospectral methods, moreover, the polynomial approximating each component of the state will be expressed as a Lagrange interpolating polynomial constructed from a Lagrange basis with $B$ time nodes. While in the Lobatto case these $B$ nodes coincide with the $N$ collocation points, in the Gauss and Radau cases they include one of the bounds of the time interval that is not a collocation point, so $B=N+1$.

%that passes through $B$ points $(t_k, x(t_k))$ for $k \in [1...B]$. The $t_k$ time points are called \textit{node points}. A polynomial constructed over $B$ node points will have $B-1$ degree and therefore $B$ independent parameters.
%\textit{Collocation points} are those where the dynamics equation is imposed. For each scheme we call $N$ the number of collocation points. Applying this condition also means that each polynomial is subject to $N$ constraints.

 %Considering whether the interval extremes are collocation points, we can classify the pseudospectral methods between Gauss methods, when neither extreme is a collocation point, Radau methods, when only one of the extremes is a collocation point, and Lobatto methods, when both extremes are collocation methods. 

The usual formulation of most pseudospectral methods assumes that the system dynamics is governed by a first order ODE of the form
\begin{equation} \label{eq:f}
	\dx = \vr{f}(\vr{x}, \vr{u}, t),
\end{equation}
where $\vr{x}$ and $\vr{u}$ are the state and control vectors \cite{tedrake2022underactuated}. However, in robotics, as in mechanics in general, the evolution of the system is often determined by a second order ODE of the form
\begin{equation}\label{eq:g}
	\ddq = \vr{g}(\vr{q},\dq,\vr{u},t),
\end{equation}
where $\vr{q}$ is the configuration and $\dq$ is its time derivative. To apply a usual collocation method, therefore, the common procedure is to cast \eqref{eq:g} into \eqref{eq:f} by introducing the velocity vector $\vr{v}$ as a new variable, defining the state as $\vr{x}=(\vr{q},\vr{v})$, and adding the constraint $\vr{v}=\dot{\vr{q}}$, so \eqref{eq:g} can be written as
\begin{subequations}
	\label{eq:gv}
	\begin{empheq}[left=\empheqlbrace]{align}
		\label{eq:qdot_v}\dq &= \vr{v}, \\
		\dv &= \vr{g}(\vr{q},\vr{v},\vr{u},t).
	\end{empheq}
\end{subequations}
One drawback of this approach is that
the configuration and velocity components of the trajectory are approximated by means of independent polynomials, when they are not, so the problem is formulated with more variables and equations than actually needed.
%has the undesired consequence of defining the problems with more variables and equations than actually needed. 
This concern has been addressed specifically in the case of the LGL formulation~\cite{ross2002exploiting}, and has been implicitly avoided through the use of \eqref{eq:g} in the case of the CHL method~\cite{Elnagar1998}, but to the best of our knowledge no similar studies have been done for the LG or LGR methods.

%Most pseudospectral methods will fall in one the following two cases:
%Either all node points are also collocation points, so $B = N$, like in the Lobatto schemes, or an additional point is added to the collocation points in order to define the node points, so $B = N+1$, like in the Gauss and Radau schemes.

A more important aspect related to the formulation using \eqref{eq:gv} is whether or not the approximating polynomial obtained for a velocity component $v$ coincides with $\dot{q}$, not only at the collocation points, where this coincidence is explicitly imposed, but all along the whole time interval. In a Lobatto method, where $B = N$, the approximating polynomials are of degree at most $N-1$, and therefore have $N$ independent parameters. Since the polynomial for $v$ is constrained to coincide with that of $\dot{q}$ at $N$ points, they must necessarily be the same, since there is only one polynomial of degree at most $N-1$  satisfying $N$ conditions.
%In the first case, since $B = N$, each polynomial has $N$ independent parameters. We apply $\vr{v}=\dq$ at $N$ independent points. Therefore, the only way to fulfill all the constraints at the same time is that for each component, the velocity polynomial is the derivative of the configuration polynomial, and therefore we can ensure that $\vr{v}(t)=\dq(t)$ for all times. 
%
In contrast, in the Gauss and Radau methods where $B = N+1$, the polynomials for $v$ and $\dot{q}$ have $N+1$ independent parameters, and imposing their coincidence at $N$ points does not force the two  polynomials to be the same.
%
%so polynomials for each component of $\vr{q}$ and $\vr{v}$ remain independent and $\vr{v}(t)=\dq(t)$ is not granted over the continuous time horizon of the problem.
%
An unexpected consequence of this fact is that, since $\vr{v}(t)\neq\dq(t)$, their derivatives will also be different even at the collocation points $t_i$ (i.e., $\dv(t_i)\neq \ddq(t_i)$ despite the fact that  $\vr{v}(t_i)=\dq(t_i)$),
and, since 
\begin{equation*}
\dv(t_i)=\vr{g}(\vr{q}(t_i),\dq(t_i), \vr{u}(t_i),t_i),
\end{equation*}
this implies that 
\begin{equation*}
\ddq(t_i)\neq \vr{g}(\vr{q}(t_i),\dq(t_i), \vr{u}(t_i),t_i), 
\end{equation*}
what means that, in the case of the Gauss and Radau methods, the second order dynamic constraints in \eqref{eq:g} are not really imposed at the collocation points, when they should.

%In these cases, it is also important to note that since the polynomials are independent and have the same degree, their derivatives will be different in general, $\dv(t)\neq\ddq(t)$, even at the collocation points. Therefore, fulfilling \eqref{eq:gv} does not imply the satisfaction of $\ddq = \vr{g}(\vr{q},\dq,\vr{u},t)$, not even at the collocation points, which contributes to increasing the dynamic transcription error along the obtained trajectories \cite{betts2010practical}. Having $\vr{q}$ and $\vr{v}$ modeled as independent polynomials of a same degree also results in inconsistent configuration and velocities trajectories. If we try to execute the trajectory with a controller that tries to follow q(t) and v(t), since these are mutually incompatible, the controller will be forced to make incessant corrections trying to fulfill the impossible task to follow both trajectories at the same time.

On the other hand, the Gauss and Radau schemes have the good property of providing a unique sequence of state values $\vr{q}(t_i)$, $i=1,\ldots,N$ for any given sequence of control values $\vr{u}(t_i)$, $i=1,\ldots,N$ and initial state $\vr{q}(t_0)$, since these involve $N+1$ constraints, which is the number of independent parameters of the polynomials.
%having $B = N+1$ allows the Gauss and Radau schemes to have a unique state trajectory associated with any given control and initial state, because for each polynomial we have $N+1$ parameters, $N$ collocation constraints and 1 initial value constraint. 
However, for the Lobatto schemes, each polynomial has just $N$ parameters, so imposing 
$N$ collocation constraints plus one initial value constraint overconstrains the problem, which can only be solved if the values $\vr{u}(t_i)$, $i=1,\ldots,N$ are restricted to a certain subspace~\cite{garg2010unified}. This makes Lobatto schemes viable for optimization but not for solving initial value problems.

% Proposta del Lluís per al paràgraf final de la intro
% Proposo això per intentar un millor tancament però 
% no sé si me n'he sortit. Guardo la versió anterior 
% a baix
In this paper, we aim to propose a pseudospectral collocation method that retains the advantages of the usual Lobatto, Gauss, and Radau methods, while avoiding their shortcomings. Specifically, we present a Legendre-Gauss collocation method for second order systems with the dynamics in \eqref{eq:g} that preserves the consistency between the polynomials for \vr{v} and \vr{q}, ensures that \eqref{eq:g} is fulfilled at the collocation points, and allows the solution of initial value problems. Using well-established benchmark problems from the literature, we show that the new method produces more accurate trajectories in comparison to those of the standard LG method, without increasing substantially the computational time needed to obtain the solutions. We call the new method ``second order'', to distinguish it from the usual ``first order'' methods that only guarantee the dynamics in \eqref{eq:f} at the collocation points. Our work can be seen as a natural continuation of the one in \cite{morenom2022collocation} for the trapezoidal and Hermite-Simpson methods.

\section{Problem formulation}
\label{sec:formulation}
%%%%%%%%%%%%%%%%%%%%%%%%%%%%%%%%%%%%%%%%%%%%

Let $\vr{x} = (\vr{q},\vr{v})$ be a tuple describing the robot state, where $\vr{q} \in \R{n_q}$ is the robot configuration and $\vr{v} = \dq$. We assume the robot dynamics is given by the ODE in \eqref{eq:g}, or equivalently by \eqref{eq:f}, where $\vr{u}\in \R{n_u}$ is the control vector of motor forces and torques. Then, given an instantaneous cost function $L(\vr{x}(t),\vr{u}(t))$, a path constraint $\vr{h}(\vr{x(t)},\vr{u}(t)) \leq \vr{0}$, and a boundary constraint $\vr{b}(\vr{x}(0), \vr{x}(t_f),t_f) = \vr{0}$, the optimal control problem that we face consists in finding trajectories $\vr{x}(t)$ and $\vr{u}(t)$, and a final time $t_f>0$, that
%%%%%%%%%%%%%%%%%%%%%%%%%%%%%%%%%%%%%%%%%%%%%%%%%%%%%%%%%%%%%%%%%%%%%%%%%
% Optimization problem using optidef package
%
% Recall syntax:
%
%\begin{mini!}[2]
%	{x,y} 						% Optimization variables
%	{f(x,y)\label{eq:OPT_cost}}	% Objective function and its label
%	{\label{eq:OPT}}			% Label of the whole optimization problem
%	{}							% Optimization result
%	\addConstraint{x<}{1,\label{eq:OPT_constr1}}{}  
% 		i.e.: {LHS}{RHS\label{constraint_label}}{extra_info}
%	\addConstraint{y<}{2,\label{eq:OPT_constr2}}{}	
% 		i.e.: {LHS}{RHS\label{constraint_label}}{extra_info}
%\end{mini!}
%%%%%%%%%%%%%%%%%%%%%%%%%%%%%%%%%%%%%%%%%%%%%%%%%%%%%%%%%%%%%%%%%%%%%%%%%
\begin{mini!}[2] 
	{\vr{x}(\cdot),\vr{u}(\cdot),t_f}
	{J(\vr{x}(t),\vr{u}(t)) = \int_{0}^{t_f}L(\vr{x}(t),\vr{u}(t))\;dt}
	{\label{eq:OCP}}
	{}
	\label{eq:OCP_cost}
	%\addConstraint
	%{\vr{x}(0)=}{\vr{x}_s,\label{eq:OCP_start}}{}
	%\addConstraint
	%{\vr{x}(t_f)=}{\vr{x}_g,\label{eq:OCP_goal}}{} 
	\addConstraint
	{\vr{\dot{x}}(t)}{=\vr{f}(\vr{x}(t),\vr{u}(t),t),\quad\label{eq:OCP_dynamics}}{t\in [0,t_f],}
	\addConstraint
	{\vr{h}(\vr{x}(t),\vr{u}(t)) \le}{\vr{0},\label{eq:OCP_path}}{t\in [0,t_f],}
	\addConstraint
	{\vr{b}(\vr{x}(0), \vr{x}(t_f),t_f)}{= \vr{0}.\label{eq:OCP_boundary}}{}
	%\addConstraint
	%{t_f\ge}{0.\label{eq:OCP_time}}{}
\end{mini!}
The goal of a collocation method is to transcribe the dynamics in~\eqref{eq:OCP_dynamics} into a discrete form, so the whole problem in \eqref{eq:OCP_cost}-\eqref{eq:OCP_boundary} can be expressed as a NLP problem to be solved. While the standard LG method departs from Eq.~\eqref{eq:OCP_dynamics} to do the transcription (Section \ref{sec:pseudospectral}), our new LG method will use Eq.~\eqref{eq:g} instead (Section \ref{sec:modified}).

%
%Using pseudospectral methods to solve \eqref{eq:OCP} involves modeling the control and state variables as polynomials whose value is known at certain points $t_k$ 
%
%The transcriptions of \eqref{eq:OCP_cost} and \eqref{eq:OCP_path} are relatively straightforward and less relevant in the context of this paper. They can be done, for example, by approximating the integral in \eqref{eq:OCP_cost} using some quadrature rule, and enforcing \eqref{eq:OCP_path} for all knot points $t_k$. The transcription of \eqref{eq:OCP_dynamics}, in contrast, is substantially more involved, and will be the main subject of the rest of the paper. In particular, we will seek to construct appropriate polynomial approximations of the solutions $\vr{x}(t)$ of the ODE in \eqref{eq:OCP_dynamics} for each interval $[t_k,t_{k+1}]$. These approximations will be defined as solutions of systems of equations which, when considered together for all intervals, will form a proper transcription of \eqref{eq:OCP_dynamics} over the whole time horizon $[0,t_f]$.

%%%%%%%%%%%%%%%%%%%%%%%%%%%%%%%%%%%%%%%%%%%%
\section{The standard Legendre-Gauss method}
\label{sec:pseudospectral}
%%%%%%%%%%%%%%%%%%%%%%%%%%%%%%%%%%%%%%%%%%%%

In the Legendre-Gauss method, the collocation points are the roots of the $N$-degree Legendre polynomial, which are interior to the interval $[-1,1]$. The time domain is assumed to coincide with this interval, and a variable $\tau$ will be used as the time variable running along this interval.
It is usual to transform the time variable of the actual time domain of the problem $t \in [0,t_f]$ into $\tau \in [-1, 1]$ through the affine transformation:
\begin{align}
	\label{eq:time_transf}
	\tau = -1 + \frac{2t}{t_f}.
\end{align}
%Pseudospectral methods are based on modeling the unknown functions $\vr{x}(t)$ and $\vr{u}(t)$ as polynomials of a certain degree.
The state and control trajectories are approximated by Lagrange polynomials of degree $N$ whose nodes are the $N$ collocation points $\tau_1,\ldots,\tau_N$ together with the initial point $\tau_0=-1$.
% For a method with $B$ node points, t
The $j^{th}$ component of the state is modeled as:
\begin{align}
	\label{eq:lagrange_sum_x}
	x^{N+1}_j(\tau) = \sum^{N}_{i=0} x_{ij}L^{N+1}_i(\tau),
\end{align}
where %$B=N+1$ is the number of node points, 
$x_{ij}$ is the value of the $j^{th}$ component of the state at the $i^{th}$ node point, and $L^{N+1}_i(\tau)$ is the corresponding Lagrange polynomial. %Note that the degree of the resulting polynomial is $B-1$ and therefore it has $B$ independent parameters.

Then, the dynamics, as defined in Eq.~\eqref{eq:OCP_dynamics}, are imposed at the $N$ collocation points.
%on some or all of these node points, depending on the method, making them collocation points.
Usually, the controls are modeled as Lagrange polynomials based on these collocation points, so
%for $N$ collocation points
the $j^{th}$ component of the control is modeled as:
\begin{align}
	\label{eq:lagrange_sum_u}
	u^N_j(\tau) = \sum^{N}_{i=1} u_{ij}L^N_i(\tau),
\end{align}
where $u_{ij}$ represents the value of the $j^{th}$ component of the control at the $i^{th}$ collocation point and $L^N_i(\tau)$ is the corresponding Lagrange polynomial.

%Then, the $N$ collocation points and $B$ node points are calculated according to the method used. 
%In the LGL and Chebyshev methods, the node points are the same as the collocation points, and therefore $B = N$. As we anticipated, this leaves us with a state defined by polynomials with $N$ independent parameters, and imposing the collocation conditions \eqref{eq:qdot_v} at the $N$ collocation points forces the polynomials of $\vr{v}$ to be the derivative of the polynomials of $\vr{q}$. In the LG and LGR methods, either the starting point or the end point is added to the collocation points, and therefore $B = N+1$. 

%To avoid repetition, let us now focus on the LG method, as the process that follows can be applied to all methods by modifying the indices accordingly. 
%We will drop the B-notation from now on, assuming that the collocation points are $\tau_1...\tau_N$, and the node points are obtained by adding the starting point $\tau_0$.

Following this definition, we can easily obtain the derivatives of $\vr{x}(\tau)$ with respect to $\tau$ through the use of a differentiation matrix. That is, for the $j^{th}$ component of the state we have
\begin{align}
	\label{eq:lagr_diff_x}
	\dot{x}^{N+1}_j(\tau) = \sum^{N}_{i=0} x_{ij}\dot{L}^{N+1}_i(\tau),
\end{align}
and, therefore, at the $k^{th}$ collocation point
\begin{align}
	\label{eq:lagr_diff_x_coll}
	\dot{x}^{N+1}_j(\tau_k) = \sum^{N}_{i=0} x_{ij}\dot{L}^{N+1}_i(\tau_k) = \sum^{N}_{i=0} D_{ki}x_{ij}.
\end{align}
We can define the \textit{Gauss pseudospectral differentiation matrix} $\vr{D}$, with size ${N \times (N+1)}$, where each element $D_{ki}$ of the matrix is the value of $\dot{L}^{N+1}_i(\tau_k)$ for $k = 1,\ldots,N$, and $i = 0,\ldots,N$, and the matrix $\vr{X}$
%of size ${(N+1) \times 2n_q}$,
whose values $X_{ij}$ are  $x_j(\tau_i)$ for \mbox{$i=0,\ldots,N$} and $j=1,\ldots,2n_q$. That is, each column of $\vr{X}$ comprises the values of the $j^{th}$ component of the state at all the node points, and each row is the value of all state components at the $i^{th}$ node point. Applying \eqref{eq:lagr_diff_x_coll} to all the collocation points of all the components of the state, we get
\begin{align}
	\label{eq:x_der_as_D_items}
	\dot{x}^{N+1}_j(\tau_k) = (\vr{DX})_{kj}
\end{align}
for $k = 1,\ldots,N$ and $j=1,\ldots,2n_q$. Note that matrix $\vr{DX}$ has size $N \times 2n_q$ while $\vr{X}$ has size $(N+1) \times 2n_q$.

Let us introduce now the matrices $\vr{X}^{LG}$,
defined as $X_{kj}=x_j(\tau_k)$ for $k = 1,\ldots,N$ and $j=1,\ldots,2n_q$ (that is, equal to $\vr{X}$ but without the first row, which corresponds to $\tau_0$, that is a node point but not a collocation point), and $\vr{U}$, 
defined as $U_{kj}=U_j(\tau_k)$ for $k = 1,\ldots,N$ and $j=1, \ldots,n_u$.
Taking also into account the difference in derivation with respect to $t$ and $\tau$, this notation allows us to finally express the enforcing of \eqref{eq:OCP_dynamics} at the collocation points as 
\begin{align}
	\label{eq:colloc_LG}
	\frac{2}{t_f}\vr{DX} = \vr{F}(\vr{X}^{LG},\vr{U})
\end{align}
where $\vr{F} = [f_1, ... , f_N]^\top$ is the discretization of $\vr{f}$ at the collocation points.
\section{A method for second order systems}
\label{sec:modified}
%%%%%%%%%%%%%%%%%%%%%%%%%%%%%%%%%%%%%%%%%%%%
As explained in Section~\ref{sec:introduction}, when the standard LG method is applied to a second order system through the application of \eqref{eq:gv}, the coincidence of the approximating polynomials for $\vr{v}$ and $\dq$ is not granted, and even worse, the second order dynamics constraint in \eqref{eq:g} is not actualpsed at the collocation points.
Our approach to overcome these problems consists in modeling only the configuration $\vr{q}(\tau)$, and not the whole state $\vr{x}(\tau)$. To do so, we will use the LG collocation points $\tau_1,\ldots,\tau_N$, but we will construct our node points by adding both the starting point $\tau_0=-1$ and the end point $\tau_{N+1}=1$, so that $B = N+2$. Therefore, our polynomials will be of degree $N+1$ and they will have $N+2$ independent parameters. This structure allows us to keep the capability to determine a state trajectory from a given initial state and a given control: for each configuration component, the polynomial has $N+2$ parameters, and is subjected to $N$ collocation constraints, plus one constraint for the initial configuration and another one for the initial speed. The velocity polynomials are simply obtained as the derivative of the configuration polynomials. 

This structure translates eq. \eqref{eq:lagrange_sum_x} for our method as:
\begin{align}
	\label{eq:lagrange_sum_LG2}
	q^{N+2}_j(\tau) = \sum^{N+1}_{i=0} q_{ij}L^{N+2}_i(\tau)
\end{align}
and similarly to \eqref{eq:lagr_diff_x}:
\begin{align}
	\label{eq:lagr_diff_x_LG2}
	\dot{q}^{N+2}_j(\tau) = \sum^{N+1}_{i=0} q_{ij}\dot{L}^{N+2}_i(\tau)
\end{align}
But now, instead of evaluating the expression only at the collocation points as in \eqref{eq:lagr_diff_x_coll}, we will use all the node points: 
\begin{align}
	\label{eq:lagr_diff_x_coll_LG2}
	\dot{q}^{N+2}_j(\tau_k) = \sum^{N+1}_{i=0} q_{ij}\dot{L}^{N+2}_i(\tau_k) = \sum^{N+1}_{i=0}D^*_{ki}q_{ij}
\end{align}
for $k = 0,\ldots,N+1$, resulting in the differentiation matrix $\vr{D}^*$ 
%of size${(N+2) \times (N+2)}$,
where each element $D^*_{ki}$ is the value of $\dot{L}^{N+2}_i(\tau_k)$ for $k = 0,\ldots,N+1$ and $i=0,\dots,N+1$, and the matrix $\vr{Q}$ 
whose values $Q_{ij}$ are  $q_j(\tau_i)$ for $i=0,\ldots,N+1$ and $j=1,\ldots,n_q$. Taking into account the relationship between $t$ and $\tau$, we can define the matrices $\vr{\dot{Q}}$ 
and $\vr{\ddot{Q}}$ of size ${(N+2) \times n_q}$, which contain the values of velocities and accelerations at the node points respectively, as:
\begin{align}
	\label{eq:deriv_matrices_LG2}
	\vr{\dot{Q}}&=\frac{2}{t_f}\vr{D^*Q}\\
	\vr{\ddot{Q}}&=\frac{2}{t_f}\vr{D^*\dot{Q}}=\left(\frac{2}{t_f}\right)^2\vr{D^*}^2\vr{Q}.
\end{align}
Similarly to as before, from each of these matrices we will define the sub-matrices $\vr{Q}^{LG}$,
%_{N \times n_q}$,
$\vr{\dot{Q}}^{LG}$,
%_{N \times n_q}$ 
and $\vr{\ddot{Q}}^{LG}$,
%_{N \times n_q}$, 
obtained by eliminating the first and last row in order to keep only the values related to the collocation points. This allows us to express the enforcing of the second order ODE \eqref{eq:g} at the collocation points as:
\begin{align}
	\label{eq:colloc_LG2}
	\vr{\ddot{Q}}^{LG} = \vr{G}(\vr{Q}^{LG},\vr{\dot{Q}}^{LG},\vr{U}),
\end{align}
where $\vr{G} = [g_1, ... , g_N]^\top$ is the discretization of $\vr{g}$ at the collocation points,
which along the usual transcription
\eqref{eq:OCP}, allows us to solve the optimization problem granting the satisfaction of \eqref{eq:g} at the collocation points and avoiding the inconsistency between $\vr{v}$ and $\dq$ along the whole time domain.

%%%%%%%%%%%%%%%%%%%%%%%%%%%%%%%%%%%%%%%%%%%%%%%
\section{Test Cases}
\label{sec:tests}
%%%%%%%%%%%%%%%%%%%%%%%%%%%%%%%%%%%%%%%%%%%%%%%

\begin{figure*}[t!]
	\begin{center}
		\includegraphics[width=1\linewidth]{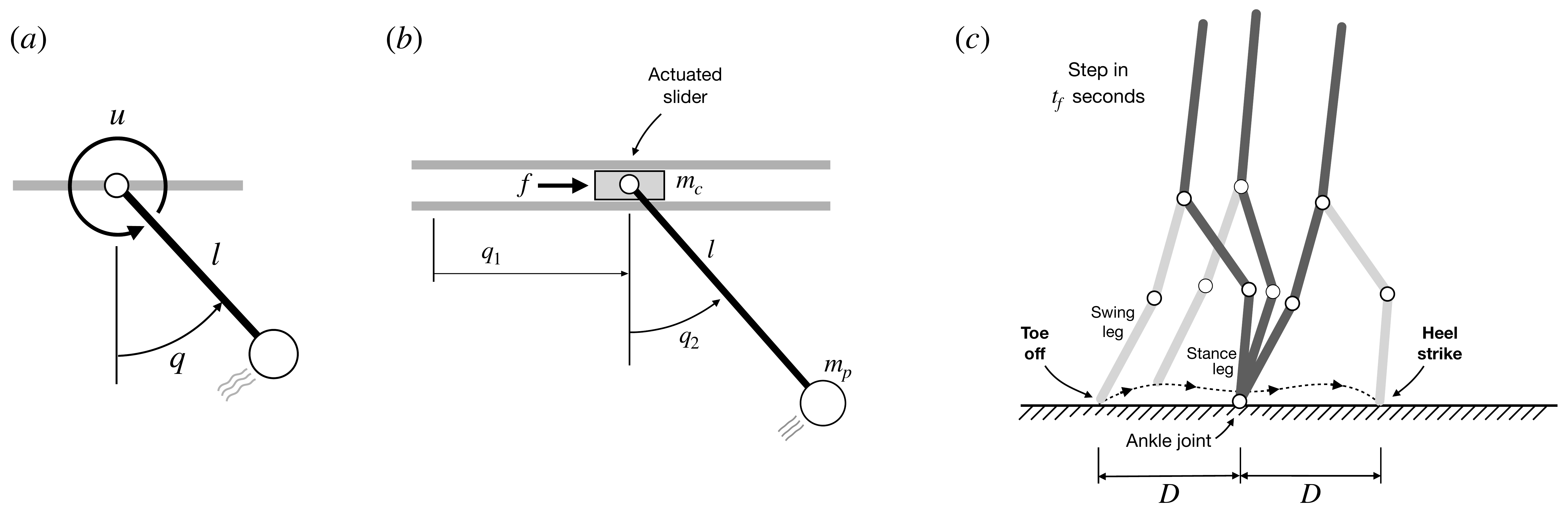}
	\end{center}
	\vspace{-2mm}
	\caption{Test cases. (a) A simple actuated pendulum. (b) A cart-pole system that has to perform a swing-up motion. (c) A five-link biped walking under a periodic gait. The three snapshots on the right illustrate the motion that occurs between the \emph{toe off} and \emph{heel strike} events defining a period of the gait.\label{fig:testcases}}
\end{figure*}

\begin{figure*}[t!]
  \begin{center}
		
		\vspace{2mm}
		\includegraphics[width=0.49\linewidth]{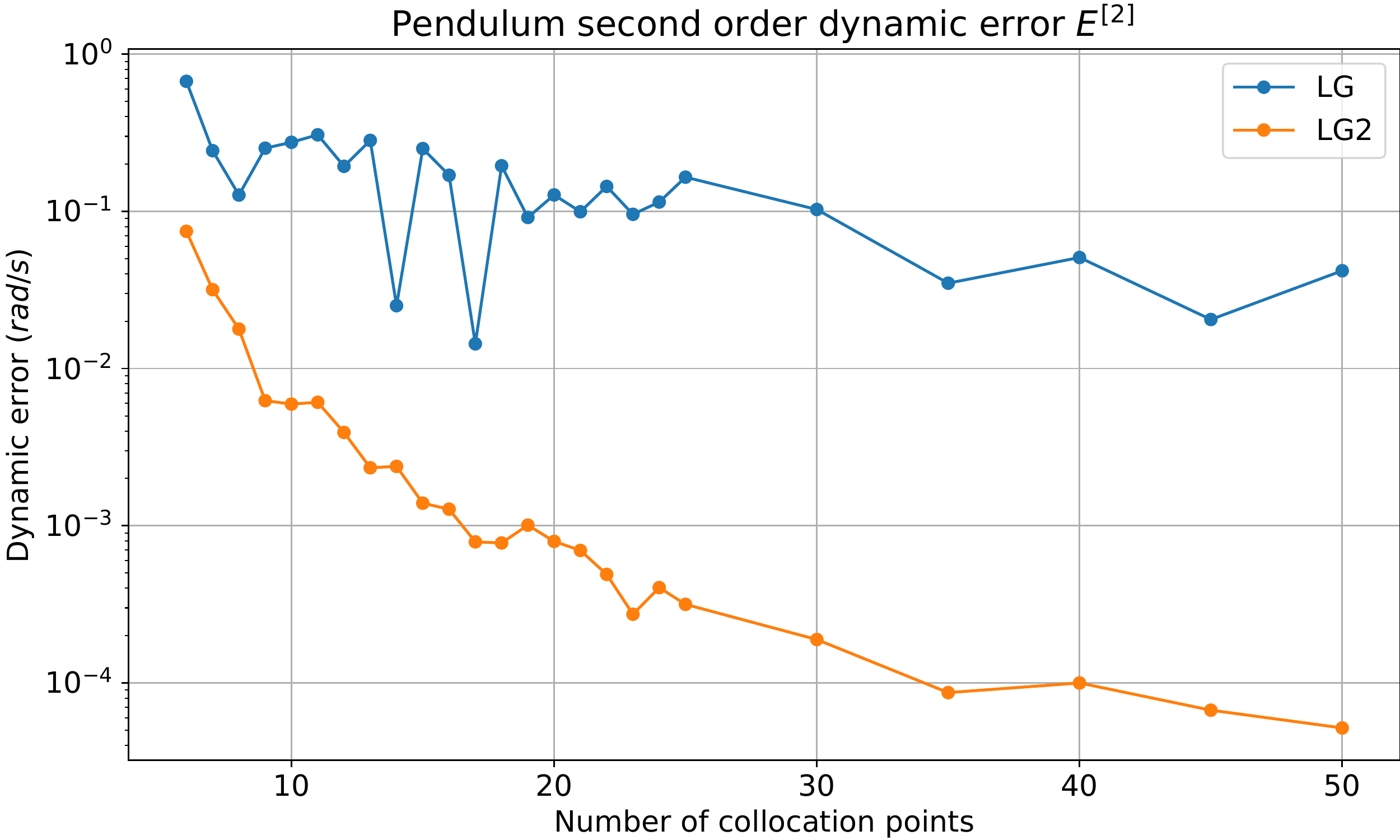}
		\includegraphics[width=0.49\linewidth]{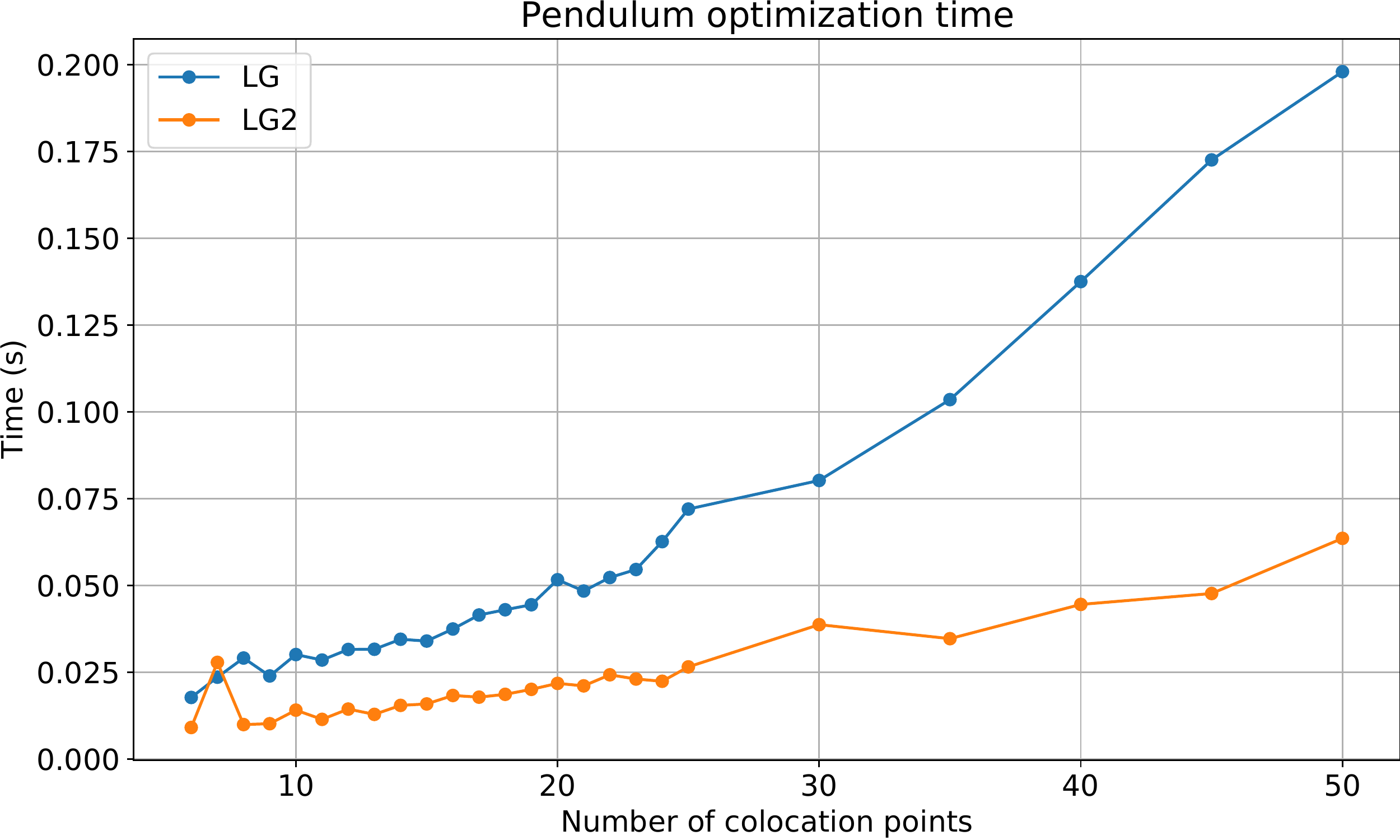}
		\vspace{2mm}
		\includegraphics[width=0.49\linewidth]{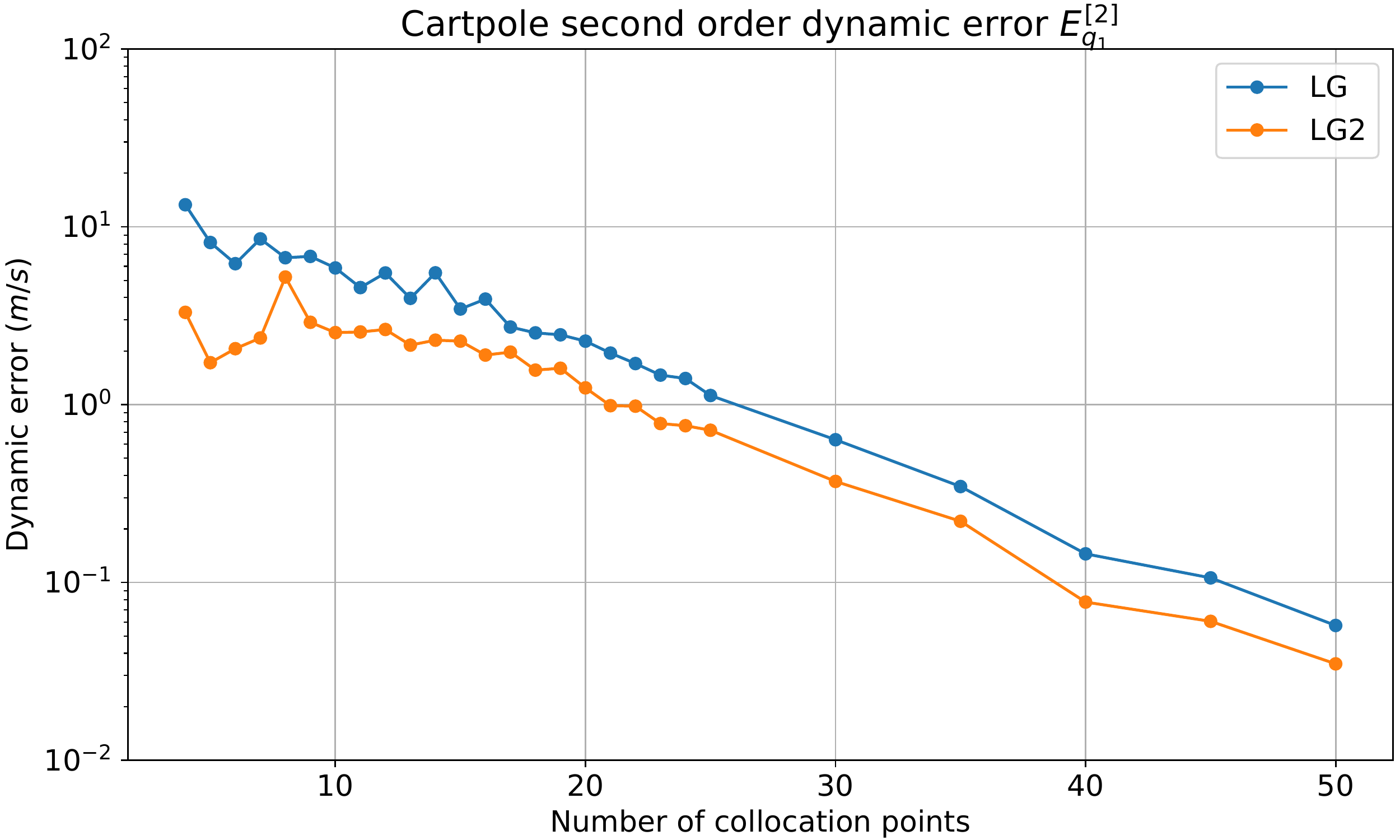}
		\includegraphics[width=0.49\linewidth]{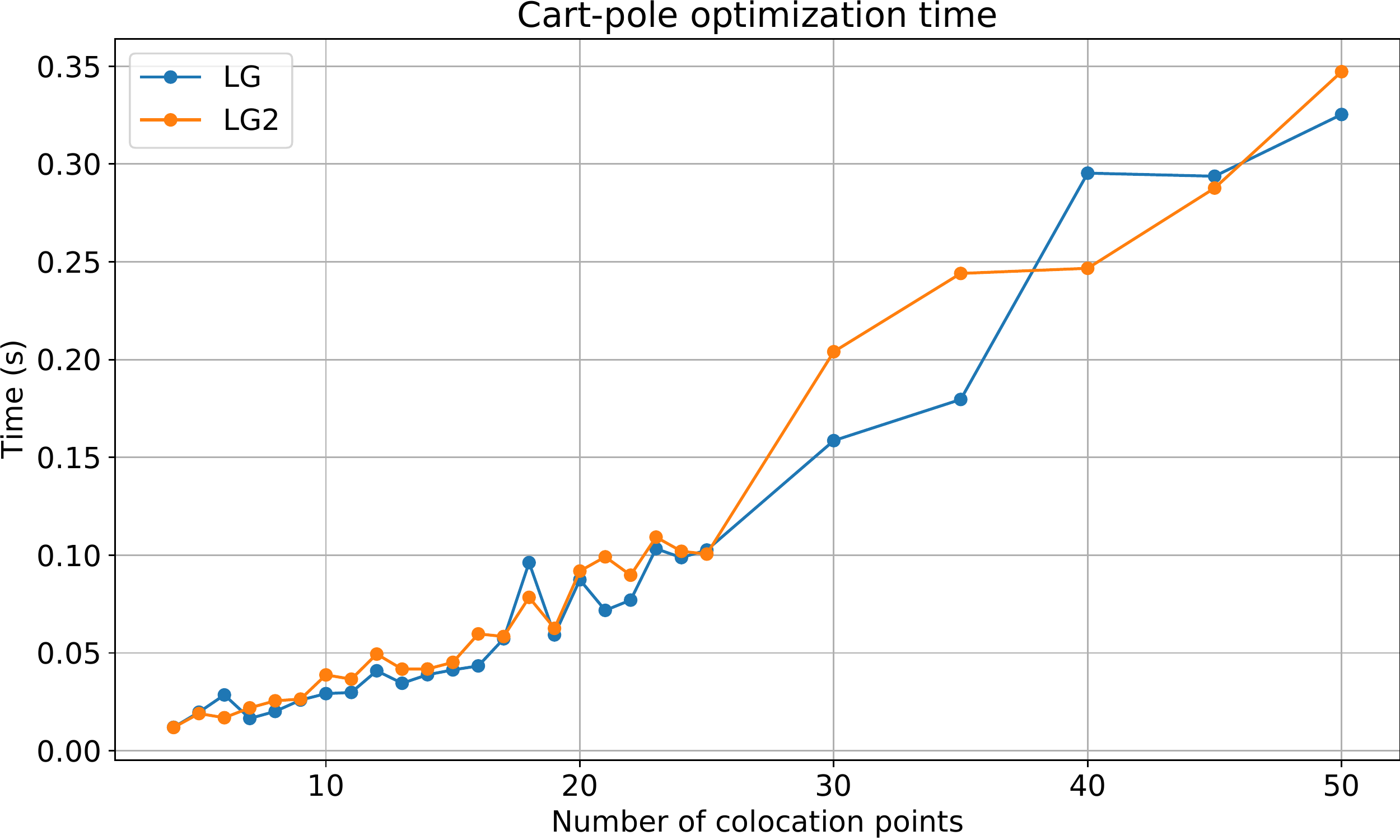}
		\vspace{2mm}
		\includegraphics[width=0.49\linewidth]{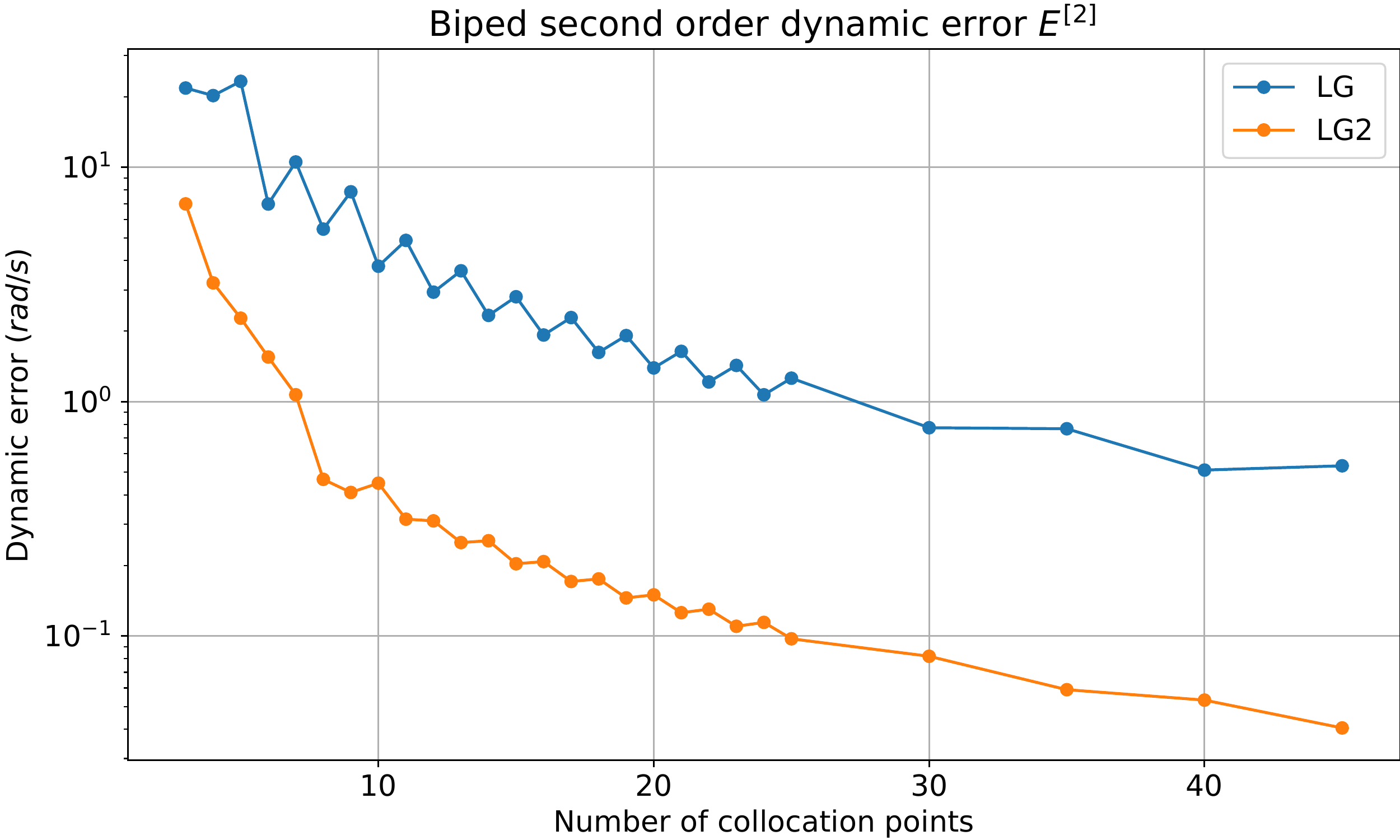}
		\includegraphics[width=0.49\linewidth]{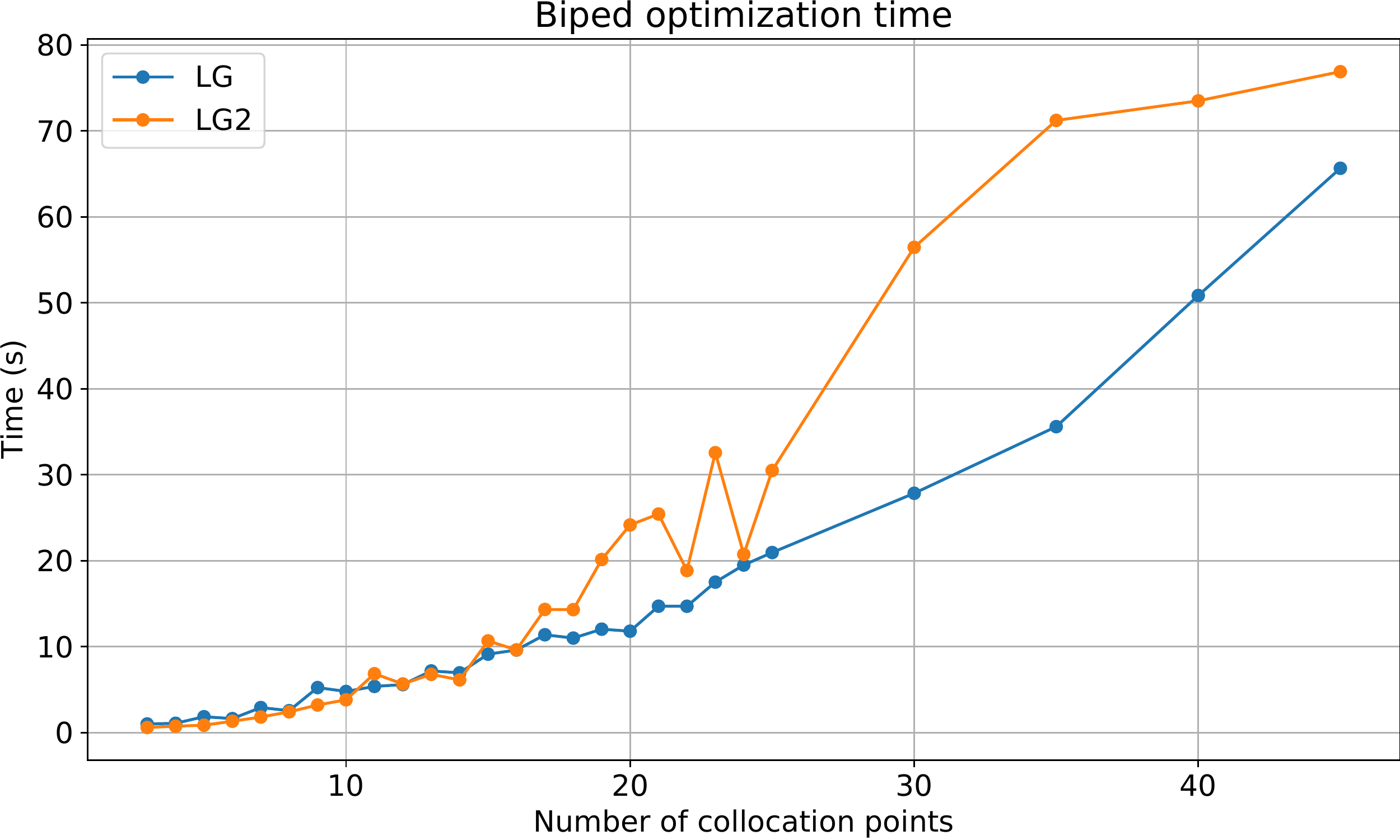}
  \end{center}
  \caption{Second order dynamic error (left plots) and time used to solve the transcribed NLP problems (right) for an increasing number of collocation points in the three test cases considered. For the cartpole system, only the plot of $E_{q_1}^{[2]}$ is provided, as the one of $E_{q_2}^{[2]}$ shows a similar trend.}
  \label{fig:results}
  %\vspace{3mm}
\end{figure*}
The first and second order versions of the Legendre-Gauss method are next compared in terms of performance, by applying them to solve the pendulum swing-up problem and two trajectory optimization problems thoroughly documented in \cite{kelly2017introduction}, namely, the cart-pole swing-up and the five-link bipedal walking problems (Fig.~\ref{fig:testcases}). Analytical solutions for these
problems are rather complex or directly not available, so in order to compare the methods, we compute the dynamic transcription error produced by each of them. To this end, we define the following errors.

The \textit{first order dynamic error} of the $q_i$ coordinate is
\begin{align}
	\label{eq:1st_order_error}
	\varepsilon^{[1]}_{q_i}(t) = \dot{q}_i(t) - v_i(t).
\end{align}
In general, this error is non-null in the first order LG method, as it does not enforce $v_i(t)=\dot{q}_i(t)$ for all $t$. However, it becomes zero by definition when using the second order method. For the same coordinate, the \textit{second order dynamic error} is
\begin{align}
	\label{eq:2nd_order_error}
	\varepsilon^{[2]}_{q_i}(t) = \ddot{q}_i(t) - g_i(\vr{q},\vr{\dot{q}},\vr{u}, t).
\end{align}
We found the \nth{2} order error more meaningful than the \nth{1} order error reported in \cite{kelly2017introduction}, since it reflects the  deviation from the actual system dynamics, which is expected to be minimized by the collocation process. When all coordinates in $\vr{q}$ have the same units, it also makes sense to define a \textit{joint error} for all coordinates. A plausible definition for this error is
\begin{align}
	\varepsilon^{[2]}(t) = 
	|\varepsilon^{[2]}_{q_1}(t)| + \ldots + |\varepsilon^{[2]}_{q_{n_q}}(t)|.
\end{align}
Finally, to summarize the error functions in just one number, we compute their integrals over $[0,t_f]$:
\begin{align}
	E^{[r]}_{q_i} &= \int_{0}^{t_f} |\varepsilon^{[r]}_{q_i}(t)| \; dt, & r=1,2,\\
	E^{[r]} &= \int_{0}^{t_f} \varepsilon^{[r]}(t) \; dt, & r=1,2.
\end{align}
%\begin{align}
%	E^{[1]}_{q_i} &= \int_{0}^{t_f} |\varepsilon^{[1]}_{q_i}(t)| \; dt, \\
%	E^{[2]}_{q_i} &= \int_{0}^{t_f} |\varepsilon^{[2]}_{q_i}(t)| \; dt, \\
%	E^{[2]} &= \int_{0}^{t_f} \varepsilon^{[2]}(t) \; dt.
%\end{align}
%

To perform the comparisons, we have implemented the methods in Python using the symbolic package SymPy \cite{meurer2017sympy} to model the systems, and the toolbox CasADi \cite{andersson2019casadi} to solve the NLP problems that result.  CasADi provides the necessary means to formulate such problems and to compute the gradients and Hessians of the transcribed equations using automatic differentiation. These are necessary to solve the NLP problems, a task for which we rely on the interior-point solver IPOPT \cite{wachter2006ipopt} in conjunction with the linear solver MUMPS \cite{amestoy2001mumps}.
%The reader can interactively reproduce our results through the online Jupyter notebooks\cite{authors2022jupytercartpole,authors2022jupyterbipedal}. 
The execution times we report have been obtained on a single-thread implementation running on an iMac computer with an Intel i7, 8-core 10th generation processor at 3.8 GHz.

In what follows, and in order to simplify the explanations, we use the shorthands LG and LG2, respectively, for the first and second order versions of the Legendre-Gauss pseudospectral methods. Fig.~\ref{fig:results} provides a comparison of the second order errors and optimization times for both methods in all problems, showing mean results after 10 runs. 

%-------------------------------------------
\subsection{The pendulum swing-up problem}
\label{subsec:pendulum}
%-------------------------------------------

The system consists of a pendulum formed by a single link connected to ground with a revolute joint (Fig.~\ref{fig:testcases}(a)). The angle of the pendulum relative to the rest position is given by $q$ and the revolute joint is powered by a motor, which applies a torque $u$ to the link. Starting with the pendulum hanging at rest in its bottom position, the goal is to reach the upright configuration with zero velocity in the shortest possible time $t_f$. The cost functional to be minimized is
\begin{equation}  \label{J_pend}
	J(u(t)) = \int_{0}^{t_f} dt = t_f.
\end{equation}

From the top plots in Fig. \ref{fig:results} we observe that the use of LG2 reduces both the second order dynamic error and the computation time. The error is reduced in about two orders of magnitude, with the gain increasing with $N$, while the optimization time is approximately halved for all $N$.
%-------------------------------------------
\subsection{The cart-pole swing-up problem}
\label{subsec:cart_pole}
%-------------------------------------------

The cart-pole system comprises a cart that travels along a horizontal track and a pendulum that hangs freely from the cart (Fig.~\ref{fig:testcases}(b)). A motor drives the cart forward and backward along the track. Starting with the pendulum hanging below the cart at rest at a given position, the goal is to reach a final configuration in a given time $t_f$, with the pendulum stabilized at a point of inverted balance and the cart staying at rest at a distance $d$ from the initial position. The cost functional to be minimized is
\begin{equation}  \label{J}
	J(u(t)) = \int_{0}^{t_f}u^2(t) dt,
\end{equation}
where $u$ is the force applied to the cart, and we adopt the same dynamic equations and problem parameters as in \cite{kelly2017introduction}.

It can be observed in the middle row of Fig. \ref{fig:results} that the use of the second order method reduces the second order dynamic error in about 40\% for all values of $N$ (left plot), while using a similar computation time in both methods (right plot).

%------------------------------------------------------------
\subsection{The 5-link bipedal walking problem}
\label{subsec:rabbit}
%------------------------------------------------------------

We next apply the methods to optimize a periodic gait for the planar biped robot shown in Fig.~\ref{fig:testcases}(c). The robot involves five links pairwise connected with revolute joints, forming two legs and a torso. All joints are powered by torque motors, with the exception of the ankle joint, which is passive. Like the cart pole system, therefore, this robot is underactuated, but it is substantially more complex. The system is commonly used as a testbed when studying bipedal walking \cite{westervelt2003hybrid,yang2009framework,park2012switching,saglam2014robust}. 

For this example we use the dynamic model given in \cite{kelly2017introduction}, which matches the one in \cite{westervelt2003hybrid} with parameters corresponding to the RABBIT prototype~\cite{chevallereau2003rabbit}. We assume the robot is left-right symmetric, so we can search for a periodic gait using a single step, as opposed to a stride, which involves two steps. This means that the state and torque trajectories will be the same on each successive step.

As in \cite{kelly2017introduction}, we define $\vr{q}$ as the vector that contains the absolute angles of all links relative to ground, while $\vr{u}$ encompasses all motor torques. Also as in \cite{kelly2017introduction}, and similarly to the cart-pole problem, our goal is to find state and action trajectories $\vr{x}(t)$ and $\vr{u}(t)$ that define an optimal gait under the cost 
\begin{equation}  \label{Jbiped}
	J(\vr{u}(t)) = \int_{0}^{t_f}\vr{u}(t)\trans\vr{u}(t) \; dt.
\end{equation}
%

%Several constraints are added to ensure a feasible gait. First of all we require the gait to be periodic, so 
%\begin{equation}
%	\vr{x}_0 = \vr{f}_H(\vr{x}_f),
%	\label{eq:heelstrike}
%\end{equation}
%where $\vr{x}_0$ and $\vr{x}_f$ are the initial and final states of the robot, and $\vr{f}_H$ is the heel-strike map. The states $\vr{x}_0$ and $\vr{x}_f$ are unknown a priori, but constrained by \eqref{eq:heelstrike}, which is the particular form of the boundary constraint \eqref{eq:OCP_boundary} in this case. To construct $\vr{f}_H$ it is assumed that, at heel strike, an impulsive collision occurs that changes the joint velocities but not their angles, and that, as soon as the leading foot impacts the ground, the trailing foot loses contact with it. The collision conserves angular momentum but introduces an instantaneous drop of kinetic energy in the system \cite{kelly2017introduction}. Next, we require the robot to march at a certain speed, which is achieved by setting the final time of the period to $t_f = 0.7$s, and the length $D$ in Fig.~\ref{fig:testcases} to $0.5$m. We also constrain the vertical velocity component of the trailing foot to be positive at $t=0$, and negative when it touches the ground for $t=t_f$. Finally, we also require the swing foot to be above the ground at all times. 

For this problem, the bottom row of Fig. \ref{fig:results} shows that our modified second order LG method results in a reduction of the second order dynamic error. Such a reduction increases with $N$ until about $N = 8$, and stabilizes at about one order of magnitude from then on. However, we can observe that this comes at the cost of increasing the optimization time a bit (by a factor of at most 2 in the worst cases) for values of $N$ above $15$.
%%%%%%%%%%%%%%%%%%%%%%%%%%%%%%%%%%%%%%%%%%%%%%%%%%%%%%%%%%%%%%%%%%%%%%%%%%%%%%%%
\section{Conclusions}
\label{sec:conclusions}
%%%%%%%%%%%%%%%%%%%%%%%%%%%%%%%%%%%%%%%%%%%%%%%%%%%%%%%%%%%%%%%%%%%%%%%%%%%%%%%%
%The most popular pseudospectral methods are the Chebyshev, Legendre-Gauss, Legendre-Gauss-Radau and Legendre-Gauss-Lobatto methods. They are often used with an first-order state-space form of optimal control problems, while the real dynamics of the systems found in robotics and other mechanics contexts are very often second order. LG and LGR methods model the state as a polynomial of degree $N$, with $N$ being the number of collocation points. This leads to incompatible state trajectories and dynamic error. LGL and Chebyshev methods model the state as polynomials of degree $N-1$, so present the problem that they impose restrictions on u(t) that make them lack the capability found in LG and LGR of having a unique state associated with each arbitrary combination of initial state and control. These problems are solved with the second order method we propose, which model the configuration trajectory $\vr{q}(t)$ as a polynomial of degree $N+1$ by adding the node points $\tau_0 = -1$ and $\tau_{N+1} = 1$ to the $N$ collocation points. 

This paper has presented a modified version of the LG pseudospectral method that is able to correctly deal with the second order nature of the dynamical systems frequently arising in robotics. In comparison to the classical LG method, the one we present guarantees that the approximation polynomials for the velocity are the time derivative of the polynomials for the configuration, which results in trajectories which are more in agreement with the system dynamics. The use of an additional node point in the definition of the configuration polynomials, moreover, has allowed us to have $N+2$ independent parameters for each component of the configuration vector. When stating an initial state and control trajectory, two initial value constraints and $N$ collocation constraints were defined for each component, all of them independent, therefore retaining the ability of the usual LG method to solve initial value problems. Using three benchmark problems from the literature, moreover, we have shown that the new LG method provides trajectories with a much smaller dynamic error in comparison to the usual LG method. This implies that the obtained trajectories will be more compliant with the system dynamics, which should facilitate their tracking control in practice.

%use less control effort, understood as the amount of additional energy necessary for the controller to perform its duty. 

Points that deserve further attention include the study of problem characteristics that lead to an increase or decrease of the computing time when switching the methods, the application of the new formulation to finite element methods that use several polynomials in sequence, instead of a single global polynomial, and a detailed theoretical analysis of the second order LG method to confirm that the usual good properties of LG methods, which include symplecticity, symmetry, and the smallest possible global error for a polynomial of a given degree \cite{hairer2006geometric}, are preserved.

%%%%%%%%%%%%%%%%%%%%%%%%%%%%%%%%%%%%
% References
%%%%%%%%%%%%%%%%%%%%%%%%%%%%%%%%%%%%

\bibliographystyle{IEEEtran}
\balance
\bibliography{IEEEabrv,references}

\end{document}